# Leveraging Group Relative Policy Optimization to Advance Large Language Models in Traditional Chinese Medicine

Jiacheng Xie[1,2], Shuai Zeng[1,2], Yang Yu[1,2], Xiaoting Tang[3], Guanghui An[4], Dong Xu[1,2]*

[1] Department of Electrical Engineering and Computer Science, University of Missouri, Columbia, MO, USA;
[2] Christopher S. Bond Life Sciences Center, University of Missouri, Columbia, MO, USA;
[3] Community Health Service Center Shanghai Pudong New Area, Shanghai, China;
[4] School of Acupuncture-Moxibustion and Tuina, Shanghai University of Traditional Chinese Medicine, Shanghai, China;
*Corresponding authors

## Abstract

Traditional Chinese Medicine (TCM) presents a rich and structurally unique knowledge system that challenges conventional applications of large language models (LLMs). Although previous TCM-specific LLMs have shown progress through supervised fine-tuning, they often face limitations in alignment, data quality, and evaluation consistency. In this study, we introduce Ladder-base, the first TCM-focused LLM trained with Group Relative Policy Optimization (GRPO), a reinforcement learning method that improves reasoning and factual consistency by optimizing response selection based on intra-group comparisons. Ladder-base is built upon the Qwen2.5-7B-Instruct foundation model and trained exclusively on the textual subset of the TCM-Ladder benchmark, using 80 percent of the data for training and the remaining 20 percent split evenly between validation and test sets. Through standardized evaluation, Ladder-base demonstrates superior performance across multiple reasoning metrics when compared to both state-of-the-art general-purpose LLMs such as GPT-4, Gemini 2.5, Claude 3, and Qwen3 and domain-specific TCM models including BenTsao, HuatuoGPT2, and Zhongjing. These findings suggest that GRPO provides an effective and efficient strategy for aligning LLMs with expert-level reasoning in traditional medical domains and supports the development of trustworthy and clinically grounded TCM artificial intelligence systems.

**Keywords:** Traditional Chinese Medicine, Large Language Models, Reinforcement Learning Alignment, Group Relative Policy Optimization

## Introduction

Traditional Chinese Medicine (TCM) [1] has been an essential part of East Asian healthcare for more than two millennia, encompassing herbal pharmacotherapy, acupuncture, and other traditional practices [2]. Its clinical relevance continues in contemporary medicine, as TCM formulations have been widely used as complementary therapies during the COVID-19 pandemic and have contributed to modern drug discovery exemplified by the isolation of artemisinin, which led to the 2015 Nobel Prize in Physiology or Medicine awarded to Tu Youyou [3]. Despite its rich empirical foundation, the corpus of classical texts, case records, and diagnostic theories in TCM remains linguistically complex and structurally unstandardized. This complexity poses challenges for conventional evidence-based research and motivates the adoption of artificial intelligence for systematic interpretation and clinical decision support.

In recent years, large language models (LLMs) have fundamentally transformed natural language understanding in both general and biomedical domains [4]. The scaling of Transformer architectures to hundreds of billions of parameters and pre-training on massive corpora have produced emergent capabilities in comprehension, reasoning, and problem solving. Frontier systems such as GPT-4 [5], Gemini 2.5 [6], Claude 3 [7], and Qwen 2.5 [8] now achieve near-human performance on professional examinations and complex reasoning benchmarks, while domain-specialized variants such as Med-PaLM [9] and BioGPT [10] have extended these advances to medical applications. However, most biomedical LLMs remain focused on Western medicine. The symbolic reasoning, holistic logic, and Classical-Chinese semantics that characterize TCM are still underexplored, leaving a substantial gap between traditional medical reasoning and modern computational intelligence.

A recent advance in reinforcement learning fine-tuning, known as Group Relative Policy Optimization (GRPO) [11], provides a promising framework for aligning models with complex domain tasks. GRPO extends the principles of Proximal Policy Optimization by sampling multiple responses for each prompt and updating model parameters through relative comparison among candidates. This approach enables the model to increase the probability of higher-quality responses while avoiding reliance on an explicit value network. GRPO has achieved significant improvements in reasoning-intensive areas such as mathematics and code generation, demonstrating greater stability and data efficiency. The same principle can be adapted to TCM modeling by defining reward signals based on measurable outcomes such as diagnostic accuracy, prescription correctness, or adherence to established treatment principles.

Several TCM-oriented large language models have been developed in recent years, including HuaTuoGPT [12], Zhongjing [13], BenTsao [14], Biancang [15], and Kimi [16]. These systems adapt general-purpose architectures to Chinese medical corpora and doctor–patient dialogues, achieving encouraging results on TCM question-answering tasks. Nevertheless, they still face important limitations, including mixed use of Western biomedical data, lack of expert-verified datasets, non-standardized evaluation procedures, and the absence of reinforcement alignment methods such as GRPO.

Here, we present the first attempt to address these limitations by applying GRPO fine-tuning to a TCM-domain LLM. We introduce Ladder-base, a reinforcement-aligned TCM language model that integrates traditional diagnostic reasoning with modern optimization strategies. Building upon a strong foundation model, we curate high-quality TCM data and employ Group Relative Policy Optimization to align the model's outputs with expert-level clinical reasoning in a data-efficient manner.

The main contributions of this work are summarized as:
(1) We develop *Ladder-base*, the first GRPO-trained TCM LLM that achieves rigorous alignment with TCM expert knowledge through comparative reinforcement learning.
(2) We construct a comprehensive TCM fine-tuning pipeline that combines curated clinical texts, structured prescription data, and domain-specific feedback to enhance interpretability and reasoning depth.
(3) We systematically evaluate Ladder-base against both leading general LLMs and prior TCM-domain models, demonstrating that GRPO alignment markedly improves diagnostic accuracy, prescription reliability, and overall domain fidelity.

## Methods

### Group Relative Policy Optimization Framework

Group Relative Policy Optimization (GRPO) [17] is a variant of the well-established Proximal Policy Optimization (PPO) algorithm (shown as Figure 1). Unlike PPO [18], GRPO removes the value function and estimates the advantage in a group-relative manner. For a given question–answer pair $(q, a)$, the behavior policy $\pi_{\theta_{old}}$ samples a group of $G$ individual responses $\{o_i\}_{i=1}^{G}$. The advantage of the $i$-th response is then computed by normalizing the group-level rewards $\{R_i\}_{i=1}^{G}$:

$$\widehat{A_{i,t}} = \frac{r_i - \text{mean}(\{R_i\}_{i=1}^{G})}{\text{std}(\{R_i\}_{i=1}^{G})}$$

The use of reward models often suffers from the issue of reward hacking [19,20], where the model optimizes the proxy reward rather than the intended objective. To mitigate this, we directly use the final accuracy of a verifiable task as the outcome reward, computed as follows:

$$R = \begin{cases} 1, & \text{if } \hat{y} = y \\ 0, & \text{otherwise} \end{cases}$$

where $y$ denotes the ground-truth answer and $\hat{y}$ represents the predicted answer. The response from the policy is parsed and evaluated using the reward system that allocates one point each for correctness, proper formatting, and accurate tagging. These components are weighted in a ratio of 5:1:1 to encourage the model to prioritize producing correct answers while maintaining appropriate structure and annotation.

Similar to PPO, GRPO employs a clipped objective function, augmented with a directly imposed Kullback–Leibler (KL) divergence penalty term:

$$\mathcal{J}_{GRPO}(\theta) = \mathbb{E}\big[q \sim P(Q), \{o_i\}_{i=1}^{G} \sim \pi_{\theta_{old}}(O \mid q)\big]$$

$$\frac{1}{G}\sum_{i=1}^{G}\frac{1}{|o_i|}\sum_{t=1}^{|o_i|}\left\{min\big[r_{i,t}(\theta)\hat{A}_{i,t}, \text{clip}\big(r_{i,t}(\theta), 1-\varepsilon, 1+\varepsilon\big)\hat{A}_{i,t}\big] - \beta\mathbb{D}_{KL}\big[\pi_{\theta}||\pi_{ref}\big]\right\}$$

where

$$r_{i,t}(\theta) = \frac{\pi_\theta(o_{i,t} \mid q, o_{i,<t})}{\pi_{\theta_{old}}(o_{i,t} \mid q, o_{i,<t})}$$

GRPO first computes the mean loss within each generated sequence and subsequently averages the losses across different samples.

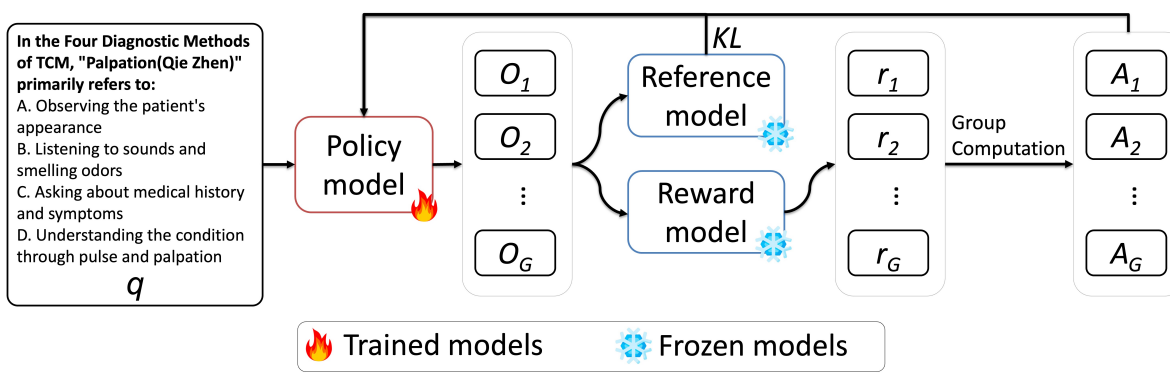

**Figure 1. Illustration of the GRPO training process.** The policy model receives a user query and generates a group of responses. Each response is assigned a reward, which is used to optimize the policy model through group-relative learning. The KL divergence term is incorporated to prevent the policy model from deviating excessively from the reference model during training.

## Training Settings

The GRPO training stage was conducted on two NVIDIA A100 PCIe GPUs (80 GB each). During the training stage, the temperature and top-p sampling parameters were set to 0.7 and 0.8, respectively. The clipping coefficient $\varepsilon$ was set to 0.2, and the weight of the KL divergence term $\beta$ was set to 0.01. Training was performed for two epochs with a group size of 6 and a batch size of 12, resulting in a total training time of approximately 60 hours. During inference, we employed greedy search to generate deterministic responses without stochastic sampling. Model training and inference were implemented using the Hugging Face Transformers library, while the GRPO process was executed via the TRL (Transformer Reinforcement Learning) framework [21]. The system prompt (as shown as Figure 2) was used to guide the model's behavior and ensure consistent output, and it was applied uniformly to all queries.

**Figure 2. Example of a query.** the text highlighted in red represents the system prompt, which guides the model's behavior and provides high-level instructions for generating responses, while the text highlighted in blue corresponds to the user query.

## Data Preparation and Preprocessing

We use the TCM-Ladder [22] benchmark from our prior work as the data source for this study. TCM-Ladder is a large-scale multimodal question-answering dataset covering major TCM domains, consisting of over 52,000 entries. It includes 21,326 high-quality QA pairs and 25,163 diagnostic dialogues derived from authoritative TCM literature and public databases, as well as herbal images and tongue images for multimodal comprehension and reasoning tasks. All data were independently verified by licensed TCM physicians to ensure accuracy and clinical consistency.

In this work, we used the textual portion of the dataset for model training. The text data were randomly split into 80% for training, 10% for validation, and 10% for testing. Based on this dataset, we further extended the previous GRPO framework to train Ladder-base, initialized from *Qwen2.5-7B-Instruct* [23], aiming to enhance the model's reasoning and robustness in text-based TCM question answering.

# Results

## Performance on diagnostic dialogue and fill-in-the-blank tasks

As shown in Figure 3, the proposed Ladder-base model was evaluated against a comprehensive set of large language models, including both general-domain systems such as GPT-4o, Gemini 2.5 Pro, and Claude 3, and TCM-domain models such as BenTsao, HuatuoGPT2, and Zhongjing. The comparison was conducted on the text-based diagnostic dialogue and fill-in-the-blank tasks from the TCM-Ladder benchmark. Six complementary metrics were used to capture linguistic accuracy and clinical reasoning quality: BLEU-4, ROUGE-L, METEOR, BERTScore, Ladder-Score, and Exact Match Accuracy.

The Ladder-base model achieved the highest overall performance, with a Ladder-Score of 0.803 and an Exact Match Accuracy of 0.8623. These values exceed those of leading general models such as GPT-4o and domain-specific counterparts such as BenTsao. Models including Gemini 2.5 Pro and Qwen3 performed competitively on general linguistic metrics but showed limited consistency in specialized reasoning. In contrast, Ladder-base, fine-tuned on the textual subset of TCM-Ladder using GRPO, demonstrated improved logical coherence and factual precision across multi-turn diagnostic dialogues. This result highlights the advantage of the GRPO-based fine-tuning framework, which optimizes model behavior through group-level reward normalization rather than individual reinforcement. The resulting model produces fluent and contextually grounded outputs, while maintaining interpretability and alignment with established TCM diagnostic reasoning. Together, these findings show that Ladder-base narrows the gap between general large language models and expert-level clinical systems for TCM.

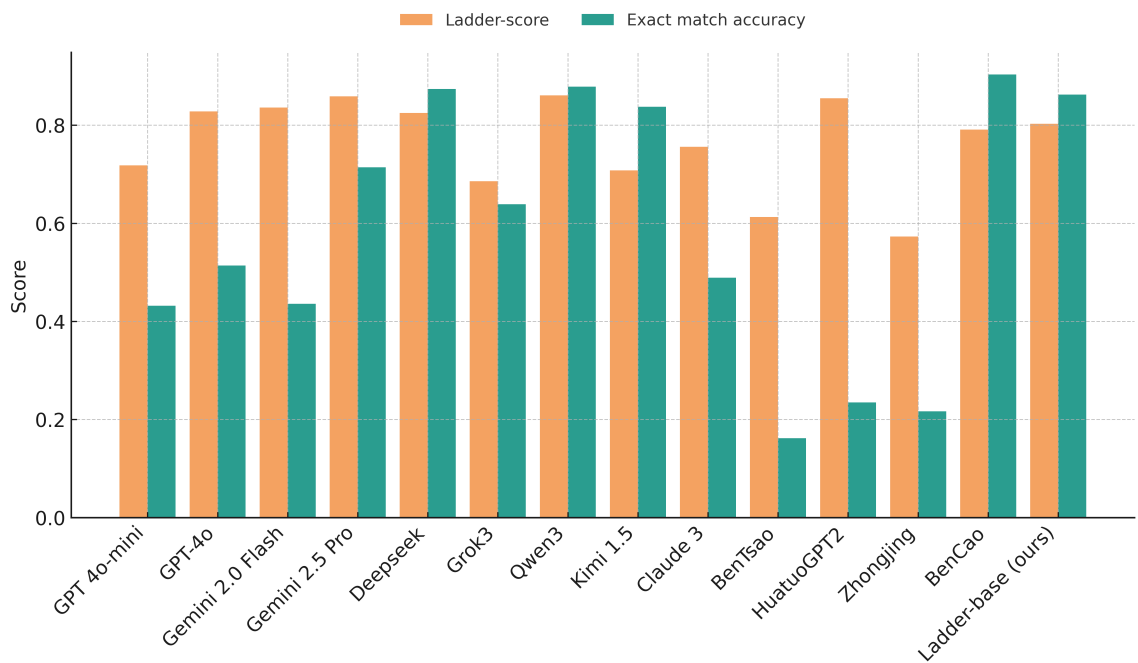

**Figure 3. Comparative performance of Ladder-base and baseline models on TCM diagnostic dialogue and fill-in-the-blank tasks.** This figure presents a comparison of 14 LLMs evaluated on text-based reasoning tasks from the TCM-Ladder benchmark. Two key metrics, Ladder-Score (orange) and Exact Match Accuracy (green), are used to assess both the quality of diagnostic dialogue generation and the precision of fill-in-the-blank responses.

## Cross-disciplinary performance on core TCM tasks

To assess model generalization across core TCM subfields, we evaluated 14 large language models on seven representative disciplines from the TCM-Ladder benchmark: Diagnostics, Pharmacognosy, Surgery, Herbal Formulas, Internal Medicine, Pediatrics, and Fundamentals. As shown in Figure 4, the proposed Ladder-base model consistently outperformed all other systems, achieving an average score of 0.7823, followed by BenCao and HuatuoGPT2, which attained 0.7241 and 0.5544, respectively. General-domain models such as GPT-4o and Gemini 2.5 Pro achieved competitive results in linguistically driven metrics but demonstrated less stability across specialized diagnostic reasoning tasks.

Performance variation across disciplines revealed that Ladder-base achieved the greatest gains in Pharmacognosy (0.9000) and Pediatrics (0.8114), suggesting improved contextual reasoning and domain adaptation when handling knowledge-intensive or symptom-dependent cases. Notably, even in disciplines characterized by ambiguous or multi-label decisions, such as Surgery and Herbal Formulas, Ladder-base maintained higher consistency than both open-domain and TCM-specific baselines.

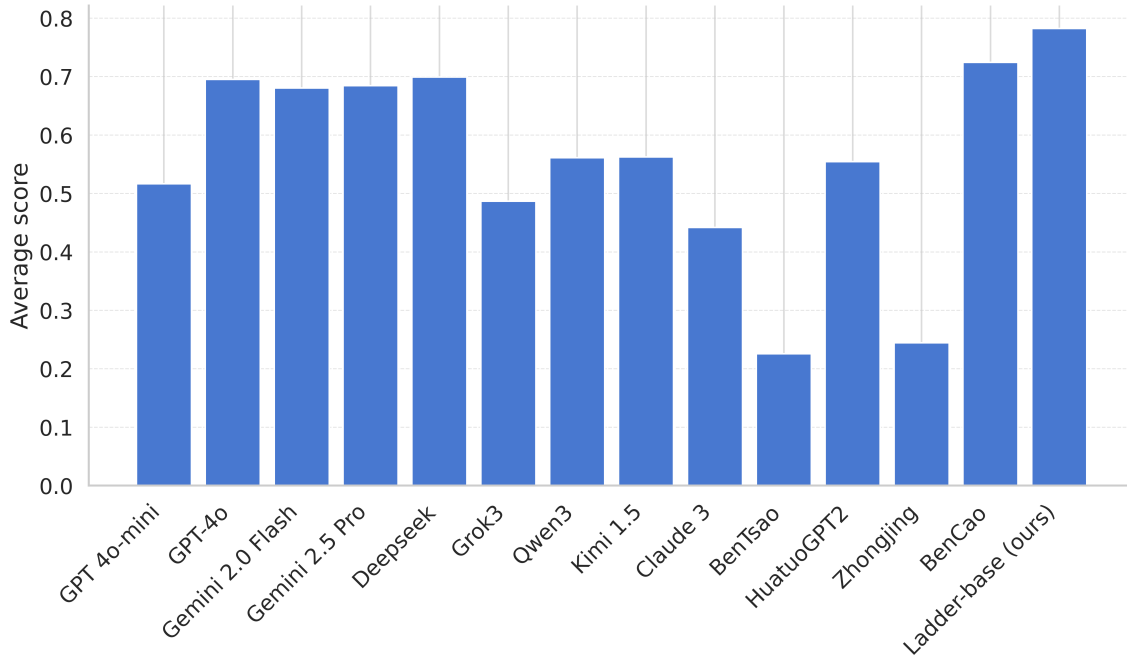

**Figure 4. Performance of large language models across TCM disciplines.** Average accuracy of 14 LLMs on seven core TCM disciplines, including Diagnostics, Pharmacognosy, Surgery, Herbal Formulas, Internal Medicine, Pediatrics, and Fundamentals.

## Discussion

The results presented in this study demonstrate that incorporating GRPO into large language model fine-tuning substantially enhances reasoning accuracy and domain robustness in TCM tasks. Compared with both general-domain and domain-specific baselines, the Ladder-base model consistently achieved superior performance in diagnostic dialogue generation and structured answer prediction. These improvements suggest that reinforcement learning with group-wise normalization provides a more stable optimization process than conventional scalar reward methods, enabling the model to better capture implicit causal patterns in clinical reasoning.

Beyond numerical gains, the enhanced reasoning ability observed in Ladder-base also reflects an important conceptual advance in adapting LLMs for knowledge systems grounded in holistic and context-dependent logic such as TCM. The integration of GRPO allows the model to learn relative judgments rather than absolute correctness, aligning more closely with how experienced physicians evaluate differential diagnoses and treatment principles. This contributes to a more interpretable and clinically consistent decision process.

Nevertheless, several limitations remain. The present work relies primarily on text-based data, which, although comprehensive, cannot fully represent the multimodal nature of TCM practice that includes visual and sensory cues. Future research should extend the GRPO-based fine-tuning framework to integrate multimodal inputs such as tongue, pulse, and herb images, as well as patient interaction data, to enable more comprehensive diagnostic reasoning. Furthermore, longitudinal studies involving expert evaluation in real clinical environments will be critical to validate the model's safety, reliability, and ethical compliance before deployment in clinical decision support systems.

## Conclusion

In summary, this study introduces a GRPO-enhanced fine-tuning framework for large language models in TCM. By training the Ladder-base model on the textual subset of the TCM-Ladder benchmark, we demonstrate that group-based reinforcement optimization can effectively improve domain reasoning, factual consistency, and response precision. The model achieves superior performance compared with both general and domain-specific baselines, highlighting the potential of GRPO to bridge the gap between general-purpose LLMs and expert-level clinical intelligence. This work provides a practical and interpretable foundation for developing trustworthy AI systems in TCM and lays the groundwork for future multimodal and clinically adaptive models.